\documentclass[conference,10pt]{IEEEtran}

\usepackage{algorithm}

\usepackage{mathtools,amsmath,amsfonts,amssymb, mathrsfs}
\usepackage{booktabs, makecell, multirow}
\usepackage{soul,xcolor,etoolbox, graphicx,tikz}
\usepackage{comment}

\usepackage[switch,columnwise]{lineno}

\usepackage{pifont}

\usepackage[utf8]{inputenc}
\usepackage[inline]{enumitem}
\usepackage[hang]{footmisc}

\usepackage[noend]{algpseudocode}

\usepackage[acronym]{glossaries}
\newacronym{ecog}{ECoG}{electrocorticogram}
\newacronym{bci}{BCI}{brain-computer interface}
\newacronym{BMI}{BMI}{brain-machine interface}
\newacronym{AI}{AI}{artificial intelligence}
\newacronym{rnn}{RNN}{recurrent neural network}
\newacronym{eeg}{EEG}{electroencephalogram}
\newacronym{snr}{SNR}{signal-to-noise ratio}
\newacronym{lstm}{LSTM}{long short-term memory}
\newacronym{svm}{SVM}{support vector machine}
\newacronym{rfc}{RFC}{random forest classifier}
\newacronym{emg}{EMG}{electromyogram}
\newacronym{psd}{PSD}{power spectral density}
\newacronym{bop}{Bop}{binary optimiser}
\newacronym{ers}{ERS}{event related synchronisation}
\newacronym{std}{STD}{standard deviation}
\newacronym{lda}{LDA}{linear discriminant analysis}
\newacronym{lightgbm}{LightGBM}{light gradient boosting machine}

\begin{document}

\setstcolor{black}

\title{Challenges and Opportunities of \\ Edge AI for Next-Generation Implantable BMIs \vspace{-5mm}}

\DeclareRobustCommand*{\IEEEauthorrefmark}[1]{\raisebox{0pt}[0pt][0pt]{\textsuperscript{\footnotesize\ensuremath{#1}}}}
\vspace{-8mm}
\vspace{3\baselineskip}

\author{
    \IEEEauthorblockN{ MohammadAli Shaeri\IEEEauthorrefmark{1}, 
                       Arshia Afzal\IEEEauthorrefmark{1,2},
                       Mahsa Shoaran\IEEEauthorrefmark{1} \\ 
    }
    \IEEEauthorblockA{
    \IEEEauthorrefmark{1}Institute of Electrical and Micro Engineering, Center for Neuroprosthetics, EPFL, 1202 Geneva, Switzerland\\
    \IEEEauthorrefmark{2}Department of Electrical Engineering, Sharif University of Technology, Tehran, Iran\\
    Email: \{mohammad.shaeri, arshia.afzal, mahsa.shoaran\}@epfl.ch
    }
\vspace{-10.5mm}
}

\maketitle

\IEEEpeerreviewmaketitle

\begin{abstract}
\small
Neuroscience and neurotechnology are currently being revolutionized by artificial intelligence (AI) and machine learning.
AI is widely used to study and interpret neural signals (analytical applications), assist people with disabilities (prosthetic applications), and treat underlying neurological symptoms (therapeutic applications).
In this brief, we will review the emerging opportunities of on-chip AI for the next-generation implantable brain machine interfaces (BMIs), with a focus on state-of-the-art prosthetic BMIs.
Major technological challenges for the effectiveness of AI models
will be discussed.
Finally, we will present algorithmic and IC design solutions to enable a new generation of AI-enhanced and high-channel-count BMIs.
\end{abstract}

\begin{IEEEkeywords}
Artificial Intelligence (AI), Machine Learning (ML), Brain Machine Interface (BMI), hardware efficiency.
\end{IEEEkeywords}
\vspace{-3mm}
\section{Introduction}
\begin{figure}
\vspace{-1mm}
  \begin{center}
    {\includegraphics[width=.36\textwidth]{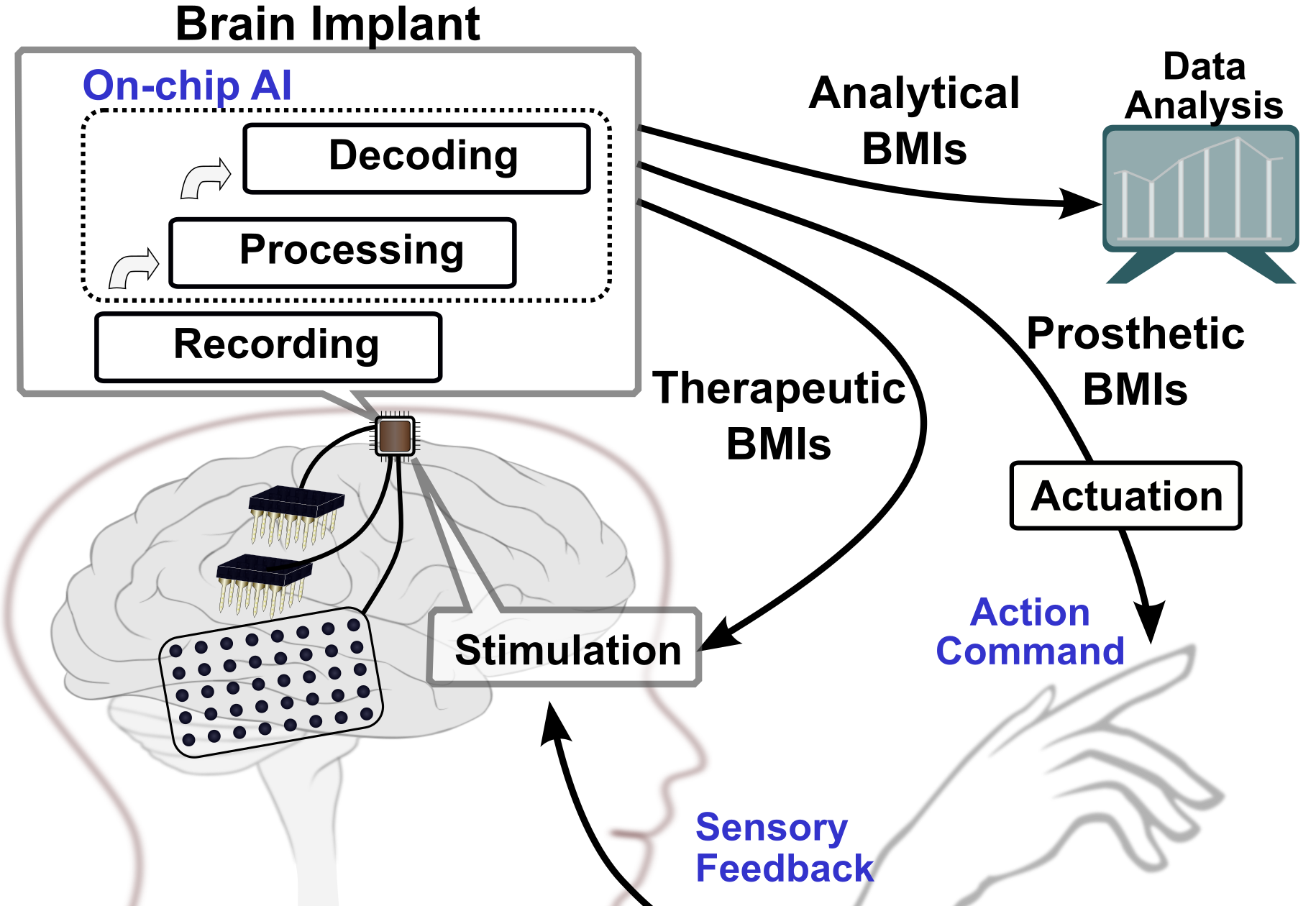}}
  \end{center}
  \vspace{-5mm}
  \caption{ \fontsize{8}{8}{ 
Schematic view of an implantable BMI. In all three BMI applications, neural signals are initially recorded and processed (e.g., amplified, digitized, filtered) and subsequently decoded (either on- or off-implant).
The output can be an actuation or neuromodulatory command for prosthetic and therapeutic applications, respectively.
}\vspace{-2mm}
}
   \label{Figure: BMI applications} 
   \vspace{-4mm}
\end{figure}

\vspace{-1mm}
\IEEEPARstart{G}{lobally}, millions of people suffer from severe motor disabilities such as paralysis and stroke. 
In order to bring disabled people back to their normal lives, a wide variety of \glspl{BMI} 
are being developed at the cutting edge of  neurotechnology and neuroscience. 
In general, \glspl{BMI} are considered as systems that close the loop from sensing to action (e.g., from vision/touch to reach/grasp), as shown in Fig.~\ref{Figure: BMI applications}. 
\
To realize this goal, an implantable BMI records 
one or more types of neural signals from the brain.
Considering the trade-off between spatiotemporal resolution and invasiveness, the intracortical and cortical recordings 
of brain activity are widely used in \Gls{BMI} applications \cite{Flesher2021Abrain, willett2021high}. 
Next, data processing and \gls{AI} techniques can be used to extract task-relevant informative content 
in the form of a movement intention or a marker of brain malfunction (e.g., brain injury).
Finally, the extracted information is used to generate an actuation command to move an artificial or natural limb, or a stimulation command to modulate the brain activity.

From the standpoint of application, BMIs can be categorized into analytical, prosthetic, and therapeutic systems (Fig. \ref{Figure: BMI applications}).
\
\textit{Analytical BMIs} are utilized to study brain activity, function, or connectivity.
Thanks to the recent success of AI in analyzing high-dimensional data, 
it is widely used in such studies ranging from cell-level (e.g., spike sorting) to cognitive-level (e.g., neural coding).
The research goal is to discover the brain mechanism or dynamics underlying 
sensory perception, or to uncover brain intention for a specific action.
\textit{Prosthetic BMIs} allow subjects to perform daily tasks such as movement \cite{Flesher2021Abrain} or typing \cite{willett2021high}.
Such BMIs can employ corticomotor activities to control natural limbs via neuromuscular 
or spinal cord stimulation. 
Another type of prosthetic BMIs stimulates the somatosensory cortex in order to restore sensory feedback \cite{Flesher2021Abrain}.
Through modulation of the nervous system, \textit{therapeutic BMIs} aim at restoring lost brain functions and treating symptoms (e.g., memory enhancement, seizure or pain suppression).

In a conventional BMI, the recorded neural signals are transmitted to an external device for further processing, while an `on-implant' AI unit performs this function in modern BMIs (Fig. \ref{Figure: BMI applications}).
Future-generation BMIs will be miniaturized and ubiquitous prostheses, applicable to daily tasks and chronic use.
Thus, designing hardware-efficient and compact implantable BMIs is of crucial importance. 
Such systems should obtain a high accuracy to be reliable for various prosthetic applications.
In this paper, we will first discuss the emerging applications and critical challenges for designing modern implantable BMIs.
Next, we will review the state-of-the-art BMI system-on-chips (SoCs) with on-chip AI. 
Finally, we will discuss novel algorithmic and circuit-level solutions to realize next-generation high-density and intelligent implantable BMIs.

\begin{table*}[t]
\caption{\vspace{-.6mm}Performance summary and comparison of the state-of-the-art BMI SoCs with on-chip AI.\vspace{-2mm}}
\label{table: Comparison} 
\vspace{-3mm}
\begin{center}
\footnotesize
\scalebox{0.8}{
\begin{tabular}{|l |m{1.7cm}|m{1.7cm}|m{1.85cm}|m{1.85cm}|m{1.85cm}|m{1.85cm}|m{1.85cm}|m{1.7cm}| }
	\hline 
	\textbf{Parameter}
		  & \textbf{\cite{boi2016bidirectional}} 
		  & \textbf{\cite{Chen2016A128}} 
		  & \textbf{\cite{Gagnon2019Wireless}} 
		  & \textbf{\cite{xu2019unsupervised}}
		  & \textbf{\cite{do2018area}}
		  & \textbf{\cite{hao202110}}
		  & \textbf{\cite{zeinolabedin202216}}
		  & \textbf{\cite{Shin2022A256}}
	\\ 
	\hline \textbf{Task} & Movement classification 
	& Movement classification 
					 &Spike sorting 
					 &Spike sorting 
					 &Spike sorting 
					 &Spike sorting 
					 &Spike sorting 
					 &Movement classification\\ 
	\hline \textbf{Input Signal} 
	& Spiking train 
	& Spiking rate 
	& Spike waveform  &Spike waveform & Spike waveform & Spike waveform & Spike waveform & ECoG\\
	\hline \textbf{Classification Model} &SNN &ELM & K-means 
	&BOTM & K-means & K-means & K-means 
	& NeuralTree \\
	\hline \textbf{Features} & -- &-- &Symmlet-2 Wavelet &-- & Filtering & Spike derivative & Adaptive \hspace{4mm} filtering & Multi-band \hspace{4mm}filters, LMP \\
	\hline \textbf{Online Training} & N & N & Y & N & N 
	& Y & Y & N \\ 
	\hline \textbf{Closed-loop} & Y & N & Y & N & N & N & N & Y \\
	\hline \textbf{\# Recording Channels} &16 &128 &10 &16  &128 &N/A 
	& 16 &256/64\\
	\hline \textbf{CMOS Process (nm)}   &180 &350 &130 &40 &65 &180 & 22 &65    \\
	\hline \textbf{Sampling Rate (kS/s)}	
					&1 
				    &0.05 
				    & N/A &24 &25 &N/A & 20 & 2      \\
	\hline \textbf{Resolution (bits)} &-- &-- 
	  & 16 &12 & 8 &-- & 9 & 10     \\
	\hline \textbf{AI Area/ch ($mm^2$)}       &3.21 &0.191$^*$ &0.08 &0.0175 &0.003 &1.023 & 0.014 & 0.0187$^\pounds$\\ 
	\hline \textbf{AI Power/ch} ($\mu$W)       &250 &0.0032$^*$ &56.9 &19     &0.175 &4.35 & 2.79  & 4.23$^\pounds$ \\ 
	\hline \textbf{System Area/ch  ($mm^2$)}     & N/A & N/A & N/A & N/A & N/A & 1.4$^\dag$
	& 0.038$^\ddag$ & 0.013$^\$$ \\ 
	\hline \textbf{System Power/ch} ($\mu$W)     & N/A & N/A & N/A & N/A & N/A & 10.25$^\dag$
	& 4.31$^\ddag$ 
	& 7.08$^\S$        \\ 

	\hline \textbf{Accuracy (\%)}             &N/A  &99.3 &N/A &93.4  &72-86 &93.2  & 94.12 & 71.5-75  \\
	\hline 
	\multicolumn{9}{l} {$^*$ Power and area  of the on-chip part (i.e, the hidden layer of ELM), divided by the number of channels.}\\
	\multicolumn{9}{l} {$^\dag$ Power and area of the analog front-end and spike detector/sorter, divided by the number of channels.}\\
	\multicolumn{9}{l} {$^\ddag$ Power and area of the analog front-end, ADC, and spike detector/sorter, divided by the number of channels.}\\
	\multicolumn{9}{l} {$^\pounds$ Power and area of the feature extractor and decoder,  divided by the number of active (i.e., selected) channels during inference (64).}\\
	\multicolumn{9}{l} {\
	\begin{tabular}{p{17cm}}
	\hspace{-3mm}$^\$$ Area of the  front-end, feature extractor, decoder, and 16-channel stimulator, divided by the  number of recording and stimulation channels (256+16).
	\end{tabular}
	}\\
	\multicolumn{9}{l} {$^\S$ Power of the mixed-signal front-end, feature extractor, and decoder, divided by the number of active channels (64).}\\
\end{tabular} 
} 
\end{center} 
\vspace{-9mm}
\end{table*}

\vspace{-1.7mm}
\section{Modern BMIs with Integrated AI \label{section:On-implant AI}}
\vspace{-1mm}
The recording capacity of implantable BMIs has grown rapidly over years, promising higher levels of proficiency and performance.
For instance, Neuralink 
and Paradromics 
develop implantable BMIs with thousands of channels.
Yet, local on-implant processing remains a challenge in many BMI directions.
Developing high-performance, energy-efficient, and scalable AI techniques could enable a new generation of implantable BMIs with minimal need for data transmission, enhanced security and privacy, and higher independence.
To achieve this goal, critical challenges at the algorithm and circuit levels must be addressed, as discussed below.
\vspace{-2mm}
\subsection{BMI Effectiveness Metrics and Design Challenges \label{subsection:BMI Effectiveness Metrics}} 
\vspace{-1.2mm}
The key dimensions for the effectiveness of AI models in the context of a BMI include
\begin{enumerate*}[label=(\Roman*)]
  \item accuracy,
  \item robustness,
  \item online adaptability, 
  \item interpretability,
  \item computational speed,
  \item scalability, and
  \item hardware efficiency. 
\end{enumerate*}

The decoding performance---often measured as classification or regression accuracy---needs to be high for clinically viable BMIs.
AI models are susceptible to performance loss due to signal variability (e.g., noise).
Therefore, the model needs to be robust to handle variations and guarantee reliability over time. 
In addition, online adaptability enables the model to adjust to non-stationary signal changes in chronic settings.
\
Another emerging AI trend is to develop interpretable models that establish an explainable relationship between brain activity and the associated physical phenomena. 
\
Furthermore, BMI systems are desired to make predictions in real-time, with minimal time required for training.
Thus, improving the training and inference speed of edge AI models is critical.

Enhancing the recording capacity of modern BMIs will significantly increase the data dimensionality.
Thus, an emerging challenge is the processing and decoding of  neural data in high-channel-count BMIs with inevitably limited hardware resources and under strict power requirements near neural tissue.
\
Thus, the hardware scalability of the model, i.e., its capacity  to handle high-dimensional data without a significant increase in hardware cost (power, chip area) is critical for implantable BMIs.
Increasing the neural data dimensionality may also increase the risk of model overfitting.
\vspace{-1.5mm}
\subsection{State-of-the-art BMI SoCs\label{subsection: Methods}} 
\vspace{-1.2mm}
AI-based methods have been utilized for spike sorting, detection of neurological symptoms (e.g., epileptic seizures \cite{zhu2021closed}), 
and motor intention decoding in a number of neural interface SoCs. 
Here, we focus on AI methods used for spike sorting and/or motor decoding in prosthetic BMIs.

\begin{figure}
  \begin{center}
  \begin{tikzpicture}
  \node[inner sep=0,scale=0.9] (pic) at (0,0)    {\includegraphics[width=.34\textwidth]{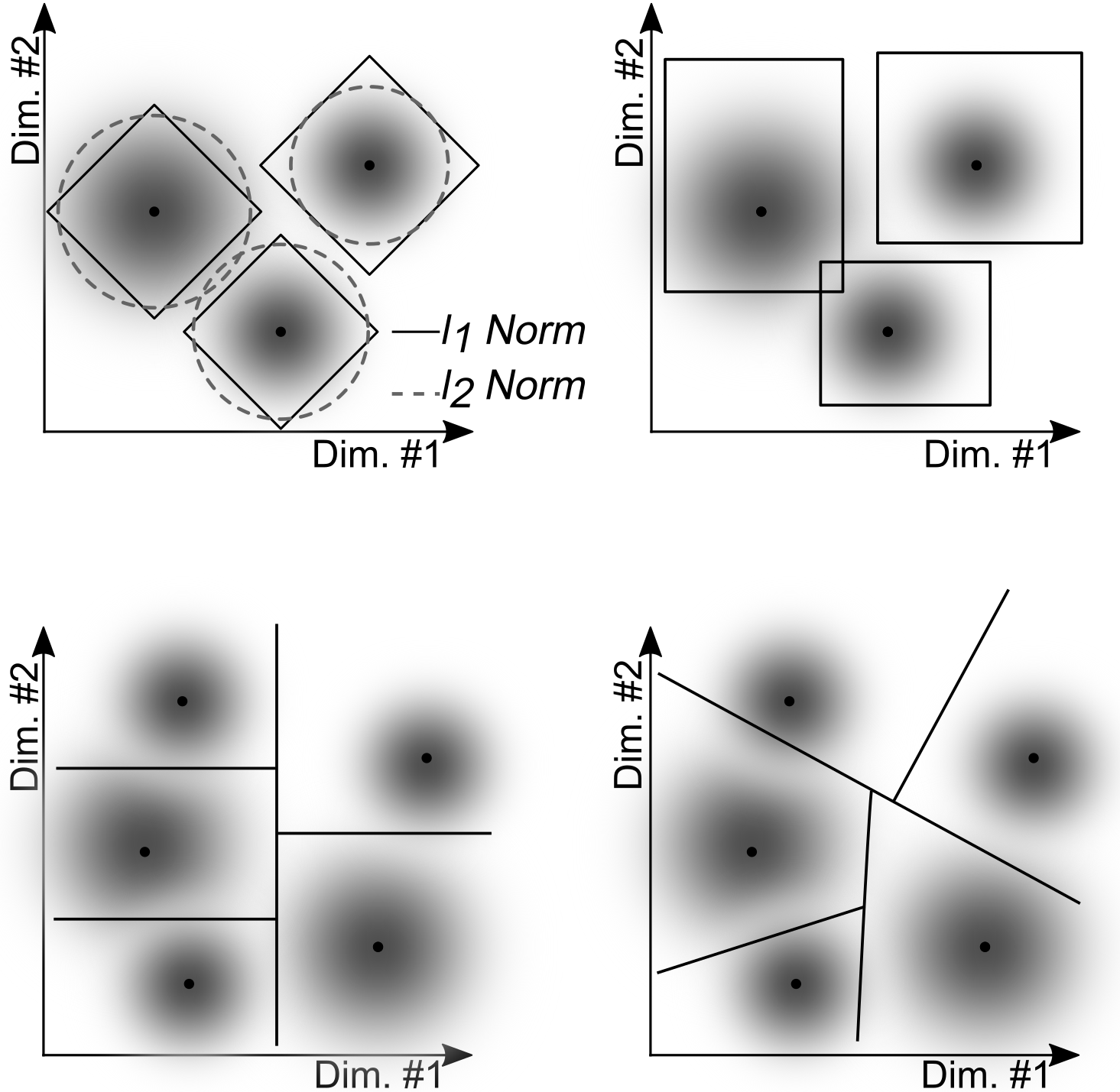}};
    \node[align=left] at (-1.75,+0.15) {(a)};
    \node[align=left] at (+1.55,+0.15) {(b)};
    \node[align=left] at (-1.75,-2.9) {(c)};
    \node[align=left] at (+1.50,-2.9) {(d)};
  \end{tikzpicture}
  \end{center}
  \vspace{-6mm}
  \caption{Hardware-efficient classification models; (a) Distance-based classification, (b) Window discrimination, (c) Axis-parallel DT (d) Oblique DT.}
   \label{Figure: Hardware-efficient classification models} 
  \vspace{-7mm}
\end{figure}

The AI models used in the BMI domain generally aim at spike sorting and movement classification. 
\
A classic approach to solve a classification/clustering problem is to allocate each data point to the neighboring class with minimal distance.
In such methods, the distance/proximity metric could be the Manhattan ($l_1$-norm) distance \cite{do2018area, Gagnon2019Wireless} 
or cosine similarity~\cite{seong2021multi}.
K-means \cite{seong2021multi, hao202110} and template matching \cite{xu2019unsupervised} are the distance-based methods widely used for spike sorting (Fig.~\ref{Figure: Hardware-efficient classification models}(a)).
\
For instance, wavelet features were extracted and used to cluster spikes based on the $l_1$-norm distance in \cite{Gagnon2019Wireless}.
The chip consumed 56.9$\mu$W/ch and occupied a silicon area of 0.08mm$^2$/ch. 
In \cite{xu2019unsupervised}, local extrema were detected from neural signal and their adjacent samples were selected and classified using the \textit{Bayes optimal template matching (BOTM)}.
This spike sorter consumed 19.0$\mu$W/ch and 0.0175mm$^2$/ch. 

An integer-coefficient filter was used for feature extraction from spikes in \cite{do2018area}, followed by feature selection and 
clustering with a simplified K-means algorithm.
Thanks to the dimensionality reduction and ultra-low-voltage SRAM usage, the fabricated spike sorter consumed 0.175$\mu$W/ch and 0.003mm$^2$/ch, albeit at the cost of a degraded accuracy (76-86\%). 
Similarly, adaptive FIR filtering 
was used for feature extraction in \cite{zeinolabedin202216}, followed by feature selection. 
Then, $l_1$-norm distance in the feature space was employed for spike sorting.
The design was fabricated in a 22nm CMOS process (2.79$\mu$W/ch and 0.014mm$^2$/ch).
\
More recently, an analog implementation of the \textit{first and second derivative extrema (FSDE)} for feature extraction 
as well as K-means were reported for spike sorting (4.35$\mu$W/ch, 1.023mm$^2$/ch)~\cite{hao202110}.

\vspace{-.5mm}
Hardware-efficient classification models such as \textit{window discrimination (WD)} have also been reported for BMIs, where a hyperrectangle discrimination window is assigned to each individual class (Fig.~\ref{Figure: Hardware-efficient classification models}(b)) \cite{Shaeri2020SFS}.
A discrimination window is composed of two decision boundaries for each data dimension, simply implemented by a few digital comparators.
\
Similarly, \textit{decision tree (DT)-based} models classify the data with a set of successive comparisons and achieve excellent energy efficiencies \cite{zhu2021closed}, Fig.~\ref{Figure: Hardware-efficient classification models}(c).
Rather than using a single feature per node as in conventional axis-parallel DTs, an \textit{oblique decision tree (OT)} uses a linear combination of features per node, thus forming a more accurate, oblique boundary for movement classification \cite{zhu2020resot} or spike sorting \cite{Yang2017AHardware} (Fig. \ref{Figure: Hardware-efficient classification models}(d)).

\textit{Neural networks-based} models have also been used in BMI applications 
\cite{boi2016bidirectional, Chen2016A128, yoo2021neural}. 
In \cite{boi2016bidirectional}, a brain-inspired \textit{spiking neural network (SNN)} was implemented to classify stimulation-evoked brain activity. 
The SNN classifier contained integrate-and-fire interconnected neurons that mimic the micro-scale property of the neuronal network.
The chip consumed 250$\mu$W/ch and 3.213mm$^2$/ch.
\
In \cite{Chen2016A128}, an \textit{extreme learning machine (ELM)} was employed to classify finger movements in monkeys. 
The ELM is a single hidden-layer feedforward network that performs a random projection of the inputs through a nonlinear hidden layer and generates the classification results through a linear output layer.
The nonlinear hidden layer was implemented on-chip, while the linear output layer was implemented off-chip via a commercial microcontroller.
The chip achieved an excellent power consumption and silicon area of 0.0032$\mu$W/ch and 0.191mm$^2$/ch, respectively. 
\
Alternatively, a \textit{binarized neural network (BNN)} with binary weights and activation function was reported as a hardware-efficient model in \cite{valencia2020neural}.
Moreover, combing a DT structure with a highly-pruned neural network led to a lightweight \textit{NeuralTree} model for finger movement classification from human ECoG \cite{Shin2022A256}.
The model comprised sparsely-connected internal nodes with fewer computations and memory needs compared to conventional OTs.
Fabricated in a 65nm process, the on-chip decoder consumed 4.23$\mu$W/ch and 0.0187mm$^2$/ch.

In addition to the classifiers discussed above, a number of AI methods seek efficient, minimal data representation to reduce hardware complexity.
\
Namely, \textit{salient feature selection} selects a small number of salient features that achieve the highest class discrimination from other classes for on-implant spike sorting \cite{Shaeri2020SFS}, thus improving the hardware efficiency.
\
Table~\ref{table: Comparison} summarizes the details of the state-of-the-art BMIs with on-chip AI.
Designs with spike sorting only, or spiking rate as input, may need additional blocks for movement classification and spiking rate extraction, respectively.

\vspace{-1.8mm}
\section{AI Algorithm and Hardware Solutions for Next-Generation BMIs  \label{section:AI Algorithm and Hardware Solutions}}
\vspace{-1mm}
Neuronal spiking rate is commonly used as the input to intracortical BMIs, requiring a complex spike sorting phase.
Recent studies, however, show that simple threshold-crossing rate (without sorting) can lead to successful decoding of motor intentions.
Although the decoding accuracy with spike sorting is generally higher, the low complexity threshold-crossing method or 
extraction of \textit{spiking band power} (i.e., the power within 0.3–1kHz) 
can be beneficial in certain applications~\cite{nason2020low-short}.

To date, various hardware-algorithm co-design solutions have been introduced to shrink the AI models and improve hardware efficiency, such as network pruning and weight quantization \cite{Yang2017AHardware, zhu2020resot}.
Similarly, SNN models represent the data with a binary stream of spiking events and facilitate hardware implementation of NNs.


A recent AI trend is to implement the training algorithm in tandem with the inference model on chip \cite{zeinolabedin202216,hao202110, shaikh2020lightweight}. 
\
Such online (i.e., adaptive) ML approaches could enable autonomous BMIs with minimal need for recalibration, at the cost of extra hardware resources required for on-chip training.

\vspace{-2mm}
\subsection{Toward Kinetic Trajectory Decoding in Implantable BMIs}
\vspace{-1.2mm}
While AI models have been implemented for discrete movement classification in BMIs, no SoC for continuous trajectory decoding has been reported so far (Table \ref{table: Comparison}).
This could be due to the  higher complexity of accurate regression models compared to binary or multi-class classifiers.
Given the crucial role of continuous trajectory decoding to control prosthetic BMIs, here we introduce an algorithmic solution that transforms the regression problem into a classification task, with key advantages over conventional regressors. 

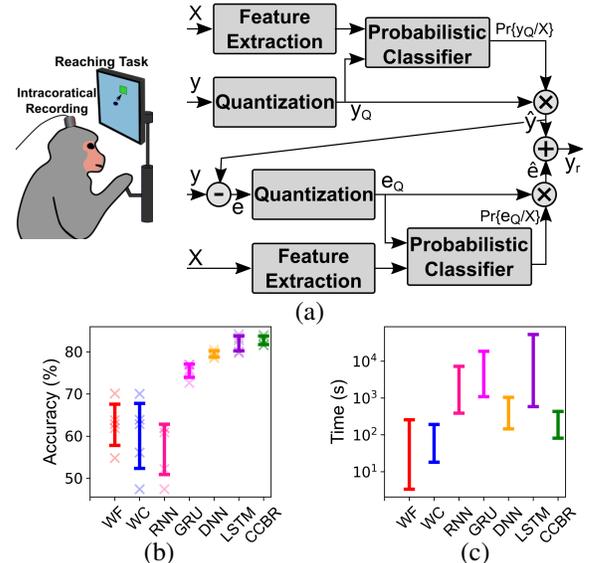
\begin{figure}[ht]
\centering
  \vspace{-4mm}
  \centering
  \begin{tikzpicture}
   \node[inner sep=0,scale=1] (pic) at (-3,0)     {\includegraphics[width=.1\textwidth]{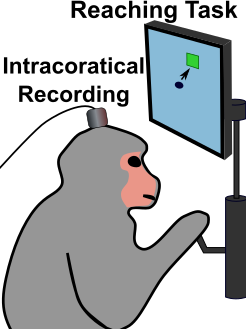}};
\vspace{-4mm}
   \node[inner sep=0,scale=1] (pic) at (1.0,0)     {\includegraphics[width=.29\textwidth]{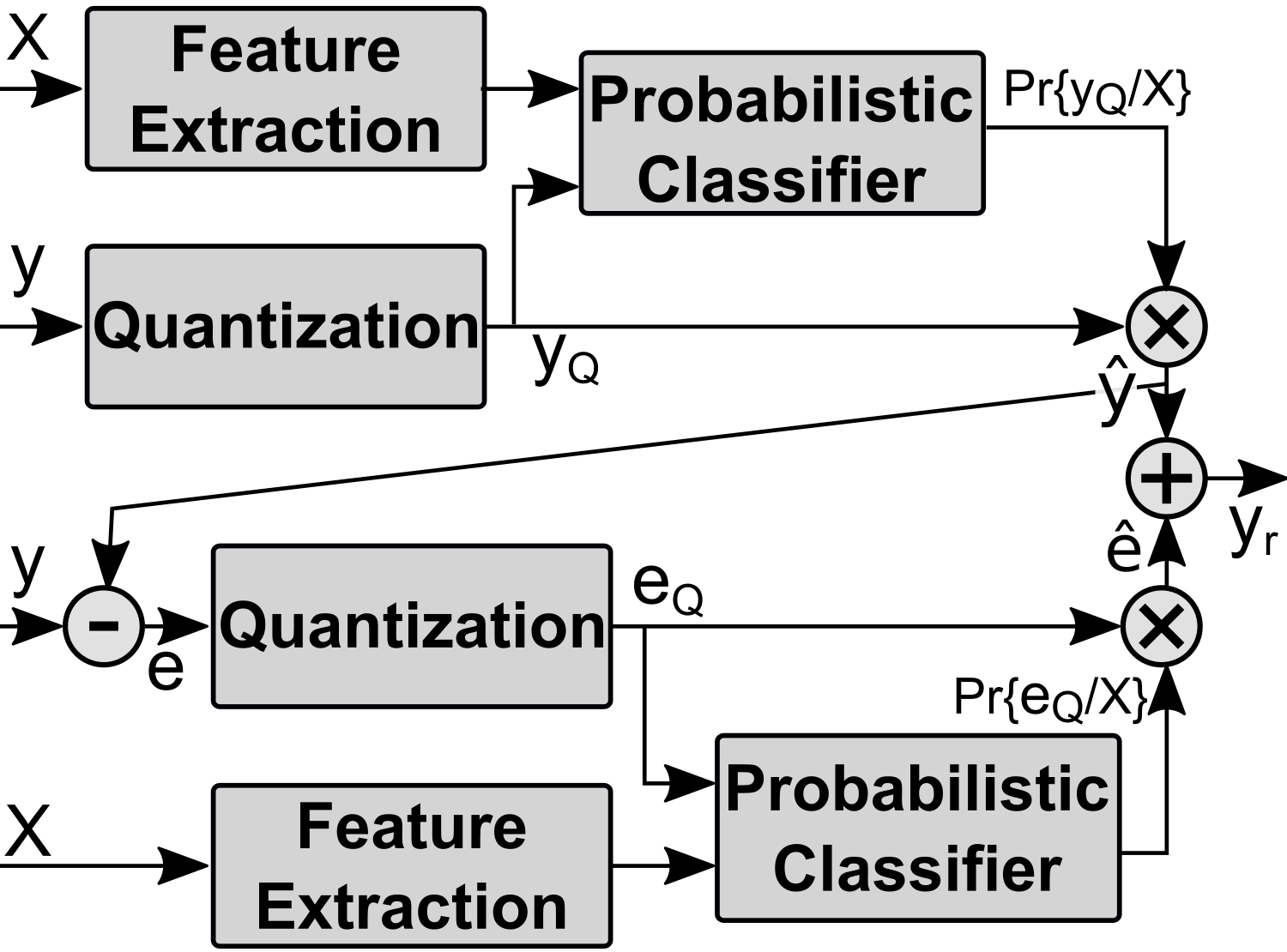}};
\vspace{-4mm}
   \node[inner sep=0,scale=1] (pic) at (0,-3.8)    {\includegraphics[width=.4\textwidth]{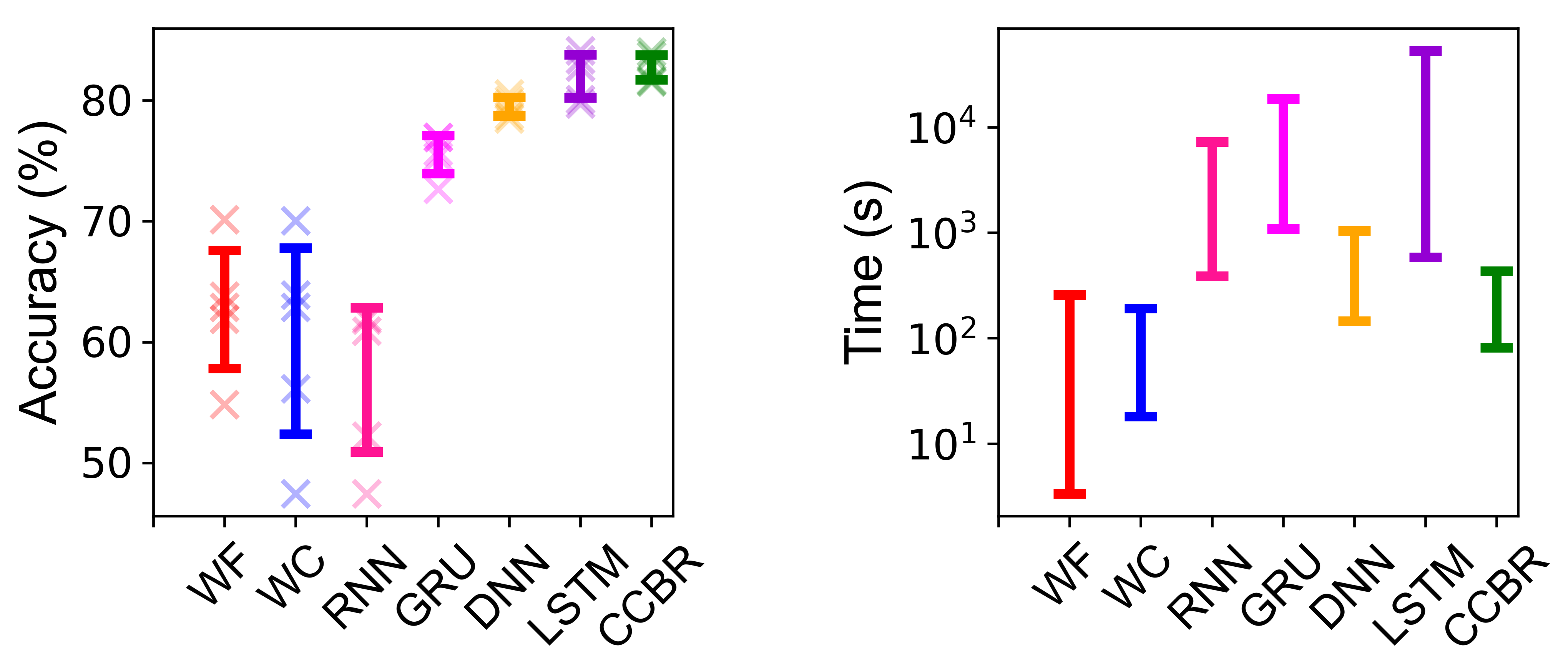}};
    \node[align=left] at (0.0,-2.2) {(a)};
    \node[align=left] at (-2.0,-5.4) {(b)};
    \node[align=left] at (+2.2,-5.4) {(c)};
  \end{tikzpicture}
    \vspace{-5mm}
\caption{
(a) Experimental setup and block diagram of the proposed CCBR for trajectory decoding.
$X$, $y$, and $e$ indicate the neural data, movement trajectory, and error signal.
(b)
Prediction accuracy (in $R^2$) for CCBR, Wiener filter (WF), Wiener cascade (WC), RNN, gated recurrent unit (GRU), and LSTM.
Error bars represent mean $\pm$ STD across validation folds.
(c) Run-time comparison for tuning different regression models vs. CCBR.
}
\label{Figure: Classification-based regression}
\vspace{-3mm}
\end{figure}
Traditionally, accurate regression models (e.g., LSTM) are used to decode movement trajectory from neural signals \cite{yaoJNE2022}. 
However, for realizing implantable BMIs, the accuracy, training speed, and hardware complexity need to be considered simultaneously.
Here, we transform the regression problem into a classification task by quantizing the kinematic signal (e.g., velocity) and solving a multi-class classification problem.
We used PCA for dimensionality reduction, 
and a simple probabilistic SVM as the classifier.
The model reconstructs the kinematic signal by multiplying the kinematic quanta (i.e., class labels) by the corresponding probabilities.
In the second phase of training, we run a similar classifier to predict the reconstruction error. 
The proposed \textit{cascaded classification-based regressor (CCBR)} generates the trajectory output by adding the predicted movement and error signals (Fig. \ref{Figure: Classification-based regression}(a)).
This model can decode the first-order along with higher-order errors and converge to maximum performance with no need for hyperparameter tuning, thus
results in fast training.


\subsubsection*{Low Sensitivity to Hyperparameters}
Reducing the sensitivity of the decoder to hyperparameters is crucial to improve the reliability and lower the need for retraining in an implantable BMI \cite{glaser2020machine}. 
\
We tested the performance of CCBR on a monkey intracortical dataset recorded from dorsal premotor cortex during a reaching task \cite{glaser2020machine}, by varying various hyperparameters, including the number of principal components (PCs), SVM's regularization factor (C), and quantization level (QL) of the movement signal ($y$). 
We observed that selecting the proper number of PCs, which indicates the input dimensionality, is sufficient to maximize the accuracy.
Thus, the model performance is highly robust to hyperparameters (C and QL).
\subsubsection*{Prediction Accuracy and Training Speed}
As shown in Fig. \ref{Figure: Classification-based regression}(b), compared to various decoders reported in \cite{glaser2020machine}, CCBR achieved the highest accuracy (82.7\%) and one of the lowest variances (1\%), proving its high reliability for neural decoding tasks.
Although complex AI models (e.g., LSTM) may result in high performance, they often require a considerable training time.
Thanks to the robustness of our proposed technique, 
we can replace the traditional methods for hyperparameter tuning (e.g., grid and random search) 
by error prediction and significantly reduce the model training time 
(Fig. \ref{Figure: Classification-based regression}(c)). 

\begin{figure}[ht]
\centering
  \vspace{-5mm}
  \centering
  \begin{tikzpicture}
   \node[inner sep=0,scale=0.9] (pic) at (0,0)     
   {\includegraphics[width=.5\textwidth]{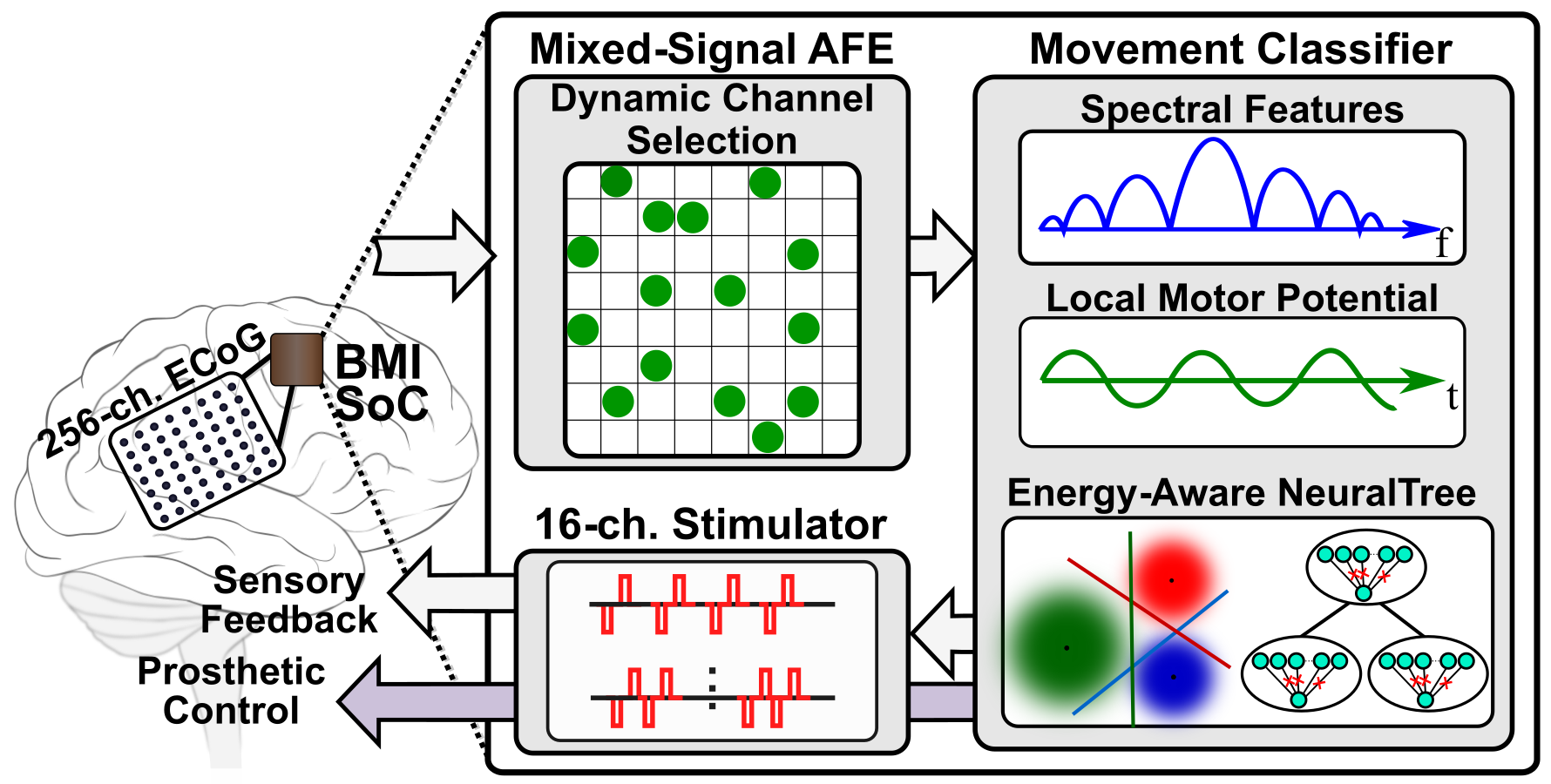}};
  \end{tikzpicture}
    \vspace{-3mm}
\caption{Block diagram of a high-density  BMI with on-chip AI.
}
\label{Figure: Implantable BMI - Block Diagram}
\vspace{-6mm}
\end{figure}

\vspace{-2.5mm}
\subsection{High-Density BMI Design with On-Chip AI} 
\vspace{-1mm}
While the existing BMI SoCs have integrated a limited number of channels (and often lack the integration of AI with the analog front-end), high-density sensing is critical to improve the motor decoding accuracy for next-generation BMIs.
One such design \cite{Shin2022A256} implements a 256-channel highly-scalable SoC for low-noise  ECoG/LFP sensing, multi-class movement classification, and  closed-loop stimulation.
The key idea is the use of a highly-pruned energy-aware AI model compatible with a dynamic channel selection scheme, and a highly-multiplexed and low-power mixed-signal front-end for successive processing of selected channels upon training.
Figure \ref{Figure: Implantable BMI - Block Diagram} depicts the architecture of this high-density BMI.
The fully-integrated SoC occupies a silicon area of 0.013mm$^2$/ch in 65nm process, with state-of-the-art power consumption and energy efficiency of 7.08$\mu$W/ch and 0.227$\mu$J/class for the entire BMI system, respectively.

\vspace{-2mm}
\section{Conclusion\label{section:Conclusion}}
\vspace{-1.5mm}
Next-Generation BMIs will be implantable prostheses with on-chip AI.
We discussed the algorithmic and hardware challenges for realizing implantable BMIs and reviewed the state-of-the-art AI models used for spike sorting and movement classification.
Finally, we introduced CCBR 
as a potential solution for trajectory decoding in Next-Generation BMIs.

\vspace{-1.5mm}
\footnotesize
\bibliographystyle{IEEEtran}
\newlength{\bibitemsep}\setlength{\bibitemsep}{-.1\baselineskip plus .02\baselineskip minus .03\baselineskip}
\newlength{\bibparskip}\setlength{\bibparskip}{0pt}
\let\oldthebibliography\thebibliography
\renewcommand\thebibliography[1]{%
  \oldthebibliography{#1}%
  \setlength{\parskip}{\bibitemsep}%
  \setlength{\itemsep}{\bibparskip}%
}
\bibliography{AICAS.bib}

\begin{thebibliography}{10}
\providecommand{\url}[1]{#1}
\csname url@samestyle\endcsname
\providecommand{\newblock}{\relax}
\providecommand{\bibinfo}[2]{#2}
\providecommand{\BIBentrySTDinterwordspacing}{\spaceskip=0pt\relax}
\providecommand{\BIBentryALTinterwordstretchfactor}{4}
\providecommand{\BIBentryALTinterwordspacing}{\spaceskip=\fontdimen2\font plus
\BIBentryALTinterwordstretchfactor\fontdimen3\font minus
  \fontdimen4\font\relax}
\providecommand{\BIBforeignlanguage}[2]{{%
\expandafter\ifx\csname l@#1\endcsname\relax
\typeout{** WARNING: IEEEtran.bst: No hyphenation pattern has been}%
\typeout{** loaded for the language `#1'. Using the pattern for}%
\typeout{** the default language instead.}%
\else
\language=\csname l@#1\endcsname
\fi
#2}}
\providecommand{\BIBdecl}{\relax}
\BIBdecl

\bibitem{Flesher2021Abrain}
S.~N. Flesher \emph{et~al.}, ``A brain-computer interface that evokes tactile
  sensations improves robotic arm control,'' \emph{Science}, vol. 372, no.
  6544, pp. 831--836, 2021.

\bibitem{willett2021high}
F.~R. Willett \emph{et~al.}, ``High-performance brain-to-text communication via
  handwriting,'' \emph{Nature}, vol. 593, no. 7858, pp. 249--254, 2021.

\bibitem{boi2016bidirectional}
F.~Boi \emph{et~al.}, ``A bidirectional brain-machine interface featuring a
  neuromorphic hardware decoder,'' \emph{Front. Neurosci.}, vol.~10, p. 563,
  2016.

\bibitem{Chen2016A128}
Y.~Chen \emph{et~al.}, ``A 128-channel extreme learning machine-based neural
  decoder for brain machine interfaces,'' \emph{IEEE Trans. Biomed. Circuits
  Syst.}, vol.~10, no.~3, pp. 679--692, June 2016.

\bibitem{Gagnon2019Wireless}
G.~{Gagnon-Turcotte} \emph{et~al.}, ``A wireless electro-optic headstage with a
  0.13-$\mu$m cmos custom integrated {DWT} neural signal decoder for
  closed-loop optogenetics,'' \emph{IEEE Trans. Biomed. Circuits Syst.},
  vol.~13, no.~5, pp. 1036--1051, Oct 2019.

\bibitem{xu2019unsupervised}
H.~Xu \emph{et~al.}, ``Unsupervised and real-time spike sorting chip for neural
  signal processing in hippocampal prosthesis,'' \emph{J. Neurosci. Meth.},
  vol. 311, pp. 111--121, 2019.

\bibitem{do2018area}
A.~T. Do \emph{et~al.}, ``An area-efficient 128-channel spike sorting processor
  for real-time neural recording with $0.175 \mu$ w/channel in 65-nm cmos,''
  \emph{IEEE Trans. {VLSI} Syst.}, vol.~27, no.~1, pp. 126--137, 2018.

\bibitem{hao202110}
H.~Hao \emph{et~al.}, ``A 10.8 $\mu$w neural signal recorder and processor with
  unsupervised analog classifier for spike sorting,'' \emph{IEEE Trans. Biomed.
  Circuits Syst.}, vol.~15, no.~2, pp. 351--364, 2021.

\bibitem{zeinolabedin202216}
S.~M.~A. Zeinolabedin \emph{et~al.}, ``A 16-channel fully configurable neural
  soc with 1.52{W/Ch} signal acquisition, 2.79{W/Ch} real-time spike
  classifier, and 1.79{TOPS/W} deep neural network accelerator in 22nm
  {FDSOI},'' \emph{IEEE Trans. Biomed. Circuits Syst.}, 2022.

\bibitem{Shin2022A256}
U.~Shin \emph{et~al.}, ``A 256-channel 0.227$\mu${J}/class versatile brain
  activity classification and closed-loop neuromodulation soc with
  0.004mm$^2$-1.51 $\mu$w/channel fast-settling highly multiplexed mixed-signal
  front-end,'' in \emph{IEEE Inter. Solid-State Circuits Conf.}, vol.~65, 2022,
  pp. 338--340.

\bibitem{zhu2021closed}
B.~Zhu \emph{et~al.}, ``Closed-loop neural prostheses with on-chip
  intelligence: A review and a low-latency machine learning model for brain
  state detection,'' \emph{IEEE Trans. Biomed. Circuits Syst.}, 2021.

\bibitem{seong2021multi}
C.~Seong \emph{et~al.}, ``A multi-channel spike sorting processor with accurate
  clustering algorithm using convolutional autoencoder,'' \emph{IEEE Trans.
  Biomed. Circuits Syst.}, 2021.

\bibitem{Shaeri2020SFS}
M.~Shaeri and A.~M. Sodagar, ``A framework for on-implant spike sorting based
  on salient feature selection,'' \emph{Nature Comm.}, vol.~11, no. 3278, pp.
  1--9, June 2020.

\bibitem{zhu2020resot}
B.~Zhu \emph{et~al.}, ``Resot: Resource-efficient oblique trees for neural
  signal classification,'' \emph{IEEE Trans. Biomed. Circuits Syst.}, vol.~14,
  no.~4, pp. 692--704, 2020.

\bibitem{Yang2017AHardware}
Y.~Yang \emph{et~al.}, ``A hardware-efficient scalable spike sorting neural
  signal processor module for implantable high-channel-count brain machine
  interfaces,'' \emph{IEEE Trans. Biomed. Circuits Syst.}, vol.~11, no.~4, pp.
  743--754, August 2017.

\bibitem{yoo2021neural}
J.~Yoo and M.~Shoaran, ``Neural interface systems with on-device computing:
  machine learning and neuromorphic architectures,'' \emph{Curr. Opin.
  Biotech.}, vol.~72, pp. 95--101, 2021.

\bibitem{valencia2020neural}
D.~Valencia and A.~Alimohammad, ``Neural spike sorting using binarized neural
  networks,'' \emph{IEEE Trans. Neur. Syst. Rehab. Eng.}, vol.~29, pp.
  206--214, 2020.

\bibitem{nason2020low-short}
S.~R. Nason \emph{et~al.}, ``A low-power band of neuronal spiking activity
  dominated by local single units improves the performance of brain--machine
  interfaces,'' \emph{Nat. Biom. Eng.}, vol.~4, no.~10, pp. 973--983, 2020.

\bibitem{shaikh2020lightweight}
S.~Shaikh \emph{et~al.}, ``Lightweight reinforcement algorithms for autonomous,
  scalable intra-cortical brain machine interfaces,'' \emph{bioRxiv}, 2020.

\bibitem{yaoJNE2022}
L.~Yao \emph{et~al.}, ``Fast and accurate decoding of finger movements from
  {ECoG} through riemannian features and modern machine learning techniques,''
  \emph{J. Neur. Eng.}, vol.~19, no.~1, pp. 1--14, Feb 2022.

\bibitem{glaser2020machine}
J.~I. Glaser \emph{et~al.}, ``Machine learning for neural decoding,''
  \emph{Eneuro}, vol.~7, no.~4, 2020.

\end{thebibliography}

\end{document}